\pdfoutput=1

\documentclass[11pt,table,xcdraw]{article}

\usepackage{acl}

\usepackage{xcolor} 
\usepackage{latexsym}
\usepackage{graphicx}
\usepackage{booktabs}
\usepackage{amsmath}
\usepackage{amsfonts}
\usepackage[varg]{txfonts} 
\usepackage{setspace} 
\usepackage{url}
\usepackage{xspace}
\usepackage{multirow}

\usepackage{wasysym}
\usepackage{textcomp}
\usepackage{gensymb}
\usepackage{tabularx}  

\definecolor{Blue}{HTML}{000099}

\usepackage[T1]{fontenc}

\usepackage[utf8]{inputenc}

\usepackage{microtype}
\usepackage{silence}  
\newcommand{\nic}{\hphantom{0}}

\title{Concept-aware Data Construction \\Improves In-context Learning of Language Models}

\author{Michal Štefánik$^{\clubsuit*}$ \and Marek Kadlčík$^\clubsuit$ \and Petr Sojka$^\clubsuit$\\
        \\$^\clubsuit$Faculty of Informatics\\
        Masaryk University, Czech Republic}

\begin{document}
\maketitle
\def\thefootnote{*}\footnotetext{Corresponding author: stefanik.m@mail.muni.cz}
\begin{abstract}

Many recent language models (LMs) are capable of \textit{in-context learning} (ICL), manifested in the LMs' ability to perform a new task solely from a natural-language instruction. Previous work curating in-context learners assumes that ICL emerges from a vast over-parametrization or the scale of multi-task training. However, recent theoretical work attributes the ICL ability to \textit{concept-dependent} training data and creates functional in-context learners even in small-scale, synthetic settings.

In this work, we practically explore this newly identified axis of ICL quality. We propose \textbf{Concept-aware Training (\textsc{CoAT})}, a framework for constructing training scenarios that make it beneficial for the LM to learn to utilize the analogical reasoning concepts from demonstrations. We find that by using \textsc{CoAT}, pre-trained transformers \textit{can} learn to better utilise new latent concepts from demonstrations and that such ability makes ICL more robust to 
the functional deficiencies of the previous models. 

Finally, we show that concept-aware in-context learners are much more effective in in-context learning a majority of unseen tasks compared to traditional instruction tuning, and fare comparably also to previous in-context learners trained in large-scale multitask learning requiring magnitudes of more training data.
\end{abstract}

\section{Introduction}
\label{sec:intro}

\begin{figure}[tbh]
    \hspace*{-3mm}
    \vspace*{5mm}
    \includegraphics[width=1.03\columnwidth]{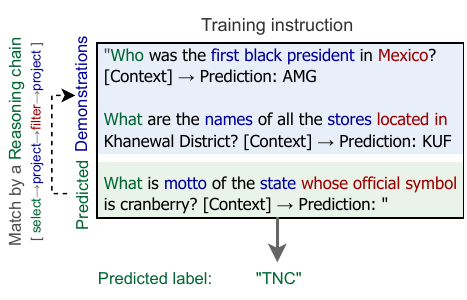}
    \vspace*{-10mm}
    \caption{%
    Example of training instruction constructed from synthetic TeaBReaC dataset where demonstrations share analogical reasoning chain. In Concept-aware Training (\textsc{CoAT}), we construct such examples to train in-context learners to utilise latent reasoning concepts whenever available in demonstrations.}
    \vspace*{-5.5mm}
    \label{fig:abstract}
\end{figure}

The in-context learning (ICL), as initially uncovered by \citet{gpt3}, is a setting requiring language models (LMs) to infer and apply correct functional relationships from the pairs of inputs and outputs (i.e.\ \textit{demonstrations}) presented in user-provided input prompt \citep{li2023transformers}. 
Given that a small set of demonstrations can be obtained for any machine learning task, in-context learning presents a much more versatile and practical alternative to training task-specific models. 

Modern in-context learners may perform ICL with quality comparable to task-specialized models \citep{Zhao2023_LLMSurvey,stefanik-etal-2023-resources}. 
However, it remains unclear why some LMs are able of ICL in such quality while others are not; Initial work introducing \textsc{GPT3} \citep{gpt3} followed by \citet{thoppilan2022lamda,chowdhery2022palm}; \textit{inter alia} explains ICL as an emergent consequence of models' scale. 
But more recent LMs \citep{sanh2022multitask,wang-etal-2022-super,flan,ouyang2022training} are based on 10 to 100 times smaller models and reach comparable ICL quality, instead attributing the ICL ability to a vast volume and diversity of pre-training tasks and instructions.

On the contrary, theoretical studies uncover different determinants of ICL quality than the model or data scale, relating ICL to specific data qualities, such as the occurrence of cases that can \textit{not} be explained by mere statistical co-occurrence of tokens. Notably, \citet{xie2022an} specify this as the occurrence of training exemplars that can \textit{only} be resolved by identifying \textbf{latent concepts}, i.e. underlying \textit{functional} relations that \textit{explain} the correct prediction. In this and other work surveyed in Section~\ref{sec:background}, authors prove that ICL can also emerge with \textit{both} small data \textit{and} small models.

Our work explores the \textit{practical} potential of concept-dependent data on the quality and robustness of in-context learning. In Section~\ref{sec:coat}, we propose and implement a data construction framework that \textit{encourages} the occurrence of concept dependencies in training data, and hence, \textit{requires} models to learn to utilise latent concepts that explain these irregularities (Fig.~\ref{fig:abstract}). We call this framework \textbf{Concept-aware Training} (\textbf{CoAT}).

In Sections~\ref{sec:experiments}, we explore the impact of \textsc{CoAT} in controlled settings. We find that (i) it is possible to train language models for in-context learning of \textit{unseen} latent concepts and (ii) that such concept-aware in-context learning \textit{is more robust} to the functional deficiencies of existing in-context learners. Finally, on a set of over 70 tasks of SuperGLUE and Natural-Instructions, we find that \textsc{CoAT} can also improve practical in-context learning performance over traditional instruction tuning approach; in many cases, \textsc{CoAT} enables ICL of otherwise not learnable tasks, and with only two training tasks reaches ICL performance \textit{comparable} to in-context learners of similar or larger size trained on massive collections of over 1,600 tasks.

\section{Background}
\label{sec:background}
\paragraph{Methods for training in-context learners}

In-context learning ability, including few-shot ICL, was first uncovered in \textsc{GPT3} \cite{gpt3} trained unsupervisedly for causal language modelling. With no other substantial differences to previous GPT models, the emergence of ICL was attributed to GPT3's \textit{scale}, having grown to over 170-billion parameters since \textsc{GPT2} ($\approx$800M params). 

Not long after, a pivotal work of \citet{pet} on a Pattern-exploiting training (PET) has shown that even much smaller (110M) models like BERT \cite{devlin-etal-2019-bert} can be fine-tuned using self-training in a similarly small data regime, first disputing the assumption on the necessity of the scale in rapidly learning new tasks.

A line of generation models further undermined the assumption of the size conditioning of ICL. Among the first, \citet{min-etal-2022-metaicl} fine-tune smaller models (<1B parameters) on a large mixture of tasks in the few-shot instructional format and find that such models can perform previously unseen tasks. Following approaches \cite{sanh2022multitask,wang-etal-2022-super} also train smaller models for instruction following on large mixtures of tasks, assuming that the model's ability to in-context learn an unseen task emerges from a large diversity of instructions and task types. A~recently popularised reinforcement learning approach of \textsc{InstructGPT} \cite{ouyang2022training} also presents an adaptation of an instruction-following objective, training on mixtures of instructions with automatic feedback.

Recently, the instruction following was extended by joint training on programming code generation \cite{chen2021evaluating} and by Chain-of-Thought (CoT) targets \cite{wei2022chain}, where the model is trained to respond with a sequence of natural-language steps deducing the answer \citep{Zhao2023_LLMSurvey,kadlcik-etal-2023-calc}. These extensions were empirically shown to enhance ICL ability \cite{fu2022gptroadmap} and were adopted by \textsc{Flan} models \cite{chung2022_flan}.

\begin{figure*}[tbh]
    \centering
    \hspace*{0mm}
    \includegraphics[width=2.1\columnwidth]{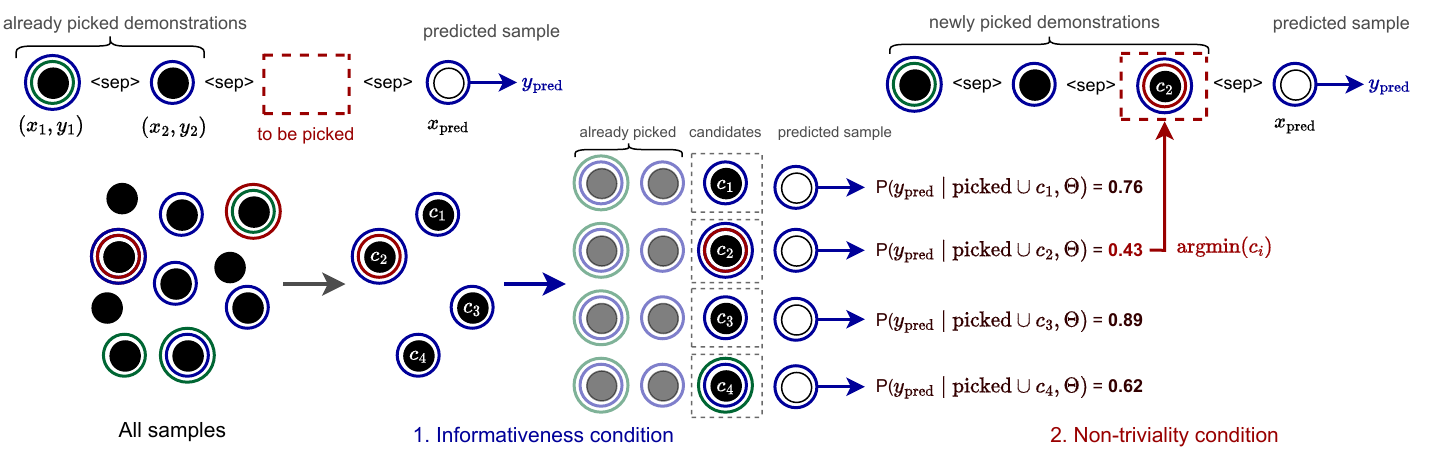}
    \caption{\textbf{Demonstrations selection in Concept-aware training (\textsc{CoAT}):} From all samples of the training dataset, we first (1)~filter out ones \textit{sharing} a specific reasoning concept {\Large \bf \textcolor{Blue} \fullmoon} with predicted sample $(x_\text{pred}, y_\text{pred})$. From this subset, we (2)~iteratively pick the candidate demonstration(s) $c_i$ such that the trained model $\Theta$'s probability of generating the correct prediction $y_\text{pred}$ if we pick $c_i$ among demonstrations is \textit{minimal}.
    }
    \label{fig:coat}
\end{figure*}

\paragraph{Analyses of ICL} 
Recent studies shed some light on the functioning of ICL in LMs through controlled experimentation, finding that the LMs' decision-making in ICL does not align with humans. 
Notably, \citet{lu-etal-2022-fantastically} report on the sensitivity of LMs to the specific formulation of the instructions in the prompt, while \citet{liu-etal-2022-makes} measures sensitivity to the ordering of in-context demonstrations. 
Further, we find that LMs perform ICL comparably well when the labels of the demonstrations are randomly shuffled \citep{what-makes-incontext-work} or when the presented CoT sequences do not make sense \citep{wang2023towards}.
We note that such behaviours \textit{differ} from learning a \textit{functional} relation from demonstrations that we expect from in-context learners \cite{li2023transformers} and can be \textit{exploited} to lead models to incorrect predictions.

Nevertheless, other studies report that under the right conditions, LMs \textit{are} able to learn functional relationships \textit{solely} from the input prompt; For instance, \citet{akyurek2023what,Li2023TransformersAA} show that Transformers can be trained to accurately learn regression functions solely from the prompt. 


\citet{xie2022an} identify the key covariate of ICL quality in the occurrence of training examples where correct predictions are conditioned by \textit{latent concepts}. 
Consider a pre-training example `Albert Einstein was [MASK]'; The correct prediction for [MASK] \textit{can} be resolved if the model can \textit{extract} and \textit{apply} a latent reasoning concept from context, such as that the context exhibits a concept of \textit{nationalities} and hence, [MASK] best substitutes `German'.
Such concept dependencies occur in language sparsely but naturally, allowing the emergence of a certain ICL quality in LMs \citep{wies2023learnability}.
Later work attributes ICL to similar data properties labelled under the terms of statistical \textit{burstiness} \cite{chan2022data} or \textit{compositionality} \citep{hahn2023theory}.


Our work builds upon this theory, but compared to previous studies limited to in-silico experiments, we elevate the idea of concept-aware training to real-world settings, with publicly available datasets and pre-trained language models. We are first to measure the impact of concept-aware data construction in \textit{extrinsic} evaluation over 70~diverse tasks and show its potential to substantially enhance data efficiency and robustness in training in-context learners, compared to previous work using magnitudes of more data and compute.






\section{Concept-Aware Training}
\label{sec:coat}

Aiming to create language models able to learn a new latent reasoning concept in-context, we propose a \textbf{Concept-Aware Training} (\textsc{CoAT}) as an instruction-tuning framework specifying \textbf{conditions for a selection of few-shot demonstrations} for the training instructions (Figure~\ref{fig:coat}).

We assume the format of training prompts widely used in the previous work training in-context few-shot learners, constructing training instructions from $k$~demonstrations consisting of the input texts $x$ with labels~$y$ followed by the predicted sample's input text $x_\text{pred}$:
\begin{equation*}
    [x_1, y_1, \langle\mathit{sep}\rangle,\ldots, x_k, y_k, \langle\mathit{sep}\rangle, x_\text{pred}] \rightarrow y_\text{pred}    
\end{equation*}
In this setting, \textsc{CoAT} proposes to filter in-context demonstrations sequentially by two conditions. The main condition, denoted as \textbf{informativeness condition}, assures to pick demonstrations exhibiting a specific \textit{reasoning concept} $\mathcal{C}$ that is \textit{shared} between a picked demonstration $(x_i, y_i)$ and the predicted example $(x_\text{pred}, y_\text{pred})$, thus picking only the demonstrations whose reasoning pattern is \textit{informative} for the correct prediction. Such settings make it beneficial for the trained model to learn to \textit{extract} and \textit{apply} concepts presented in demonstrations.


However, as the sole \textit{informativeness} condition may easily pick demonstrations very similar or identical to the predicted sample, we propose a second, \textbf{non-triviality condition}. This condition chooses from the informative demonstrations the ones with which it is `difficult' for the model to respond correctly. This condition avoids the occurrence of in-context demonstrations \textit{identical} to the predicted sample and may also increase the heterogeneity of different concepts that co-occur among the demonstrations, avoiding the over-reliance on the presence of a small set of specific concepts in small-data settings.

We note that a body of previous work proposes better ways for picking in-context demonstrations during the \textit{inference} (without updating the model) \citep{li-etal-2023-unified,gupta-etal-2023-coverage,luo2024incontext}. While some of these strategies are applicable in picking informative demonstrations in CoAT, note that this line of work is complementary to ours in assuming an existing in-context learner. More importantly, our motivation is substantially different; CoAT uses demonstrations of instruction training as a vehicle for creating training cases conditioned on latent concepts. This idea is not restricted to instruction tuning, but we note that instruction tuning presents an opportunity to implement it easily.

\paragraph{What constitutes a concept applicable in CoAT?} We broadly define the term \textit{concept} as an arbitrary functional relation of input and prediction \cite{xie2022an} that holds robustly for \textit{any} sample of a given task. Hypothetically, once the model learns to \textit{model} a specific concept perfectly, it will never produce a prediction that violates this concept \cite{stefanik-kadlcik-2023-context}.
In practice, attempts to clearly present useful concepts to the model can be obstructed by other covariates \cite{mikula2023think}, such as frequent predictive co-occurrences of tokens. Thus, we propose to pick CoAT's concepts among features that are unlikely to be substitutable by non-robust covariates, presenting as best candidates to be features conditioned on a deep or holistic decomposition of the input.

Note that the goal of CoAT is \textit{not} to represent chosen training concepts in the model's weights \textit{explicitly} but to improve the model's ability to \textit{extract} and \textit{apply} available concepts from demonstrations. Towards this goal, CoAT fundamentally assumes that the ability to extract and apply one concept transfers to \textit{other} concepts beyond the training distribution. We verify this ability later in Section~\ref{sec:rq1}.


\subsection{Proposed Implementation}

In our experiments, we implement the proposed \textsc{CoAT} framework in two training stages: First, we train LM on a scalable synthetic QA dataset, which, contrary to traditional QA datasets, contains annotations of reasoning concepts. Second, we refresh the LM's ability to work with natural language prompts by further training on a QA dataset with only natural language inputs. Hence, contrary to previous work utilising massive multitask training, in total, we only use \textit{two} QA datasets.

\paragraph{Informativeness condition}
\label{sec:coat_implementation}

We find a large collection of annotated reasoning concepts in a TeaBReaC dataset of \citet{trivedi-etal-2022-teaching}, containing more than 900 unique explanations over a large set of \textit{synthetic} QA contexts. Each TeaBReaC's explanation maps a natural question to the answer span through a sequence of declarative \textit{reasoning steps}, such as ``select$\rightarrow$group$\rightarrow$project''. Within \textsc{CoAT}, we use these explanations as the shared concepts $\mathcal{C}$ (Fig.~\ref{fig:abstract}); In the training prompts, all demonstrations exhibit the same reasoning chain as the predicted sample.

To restore the model's ability to work with a natural language, in the second step, we fit the resulting model to \textit{natural} inputs by further fine-tuning on AdversarialQA dataset \citep{bartolo-etal-2021-improving}; As the annotations of reasoning concepts in general QA datasets are scarce, in this case, we naively use the initial word of the question (``Who'', ``Where'', \ldots) as the shared concept, aware that such-grouped samples are not always mutually informative.

\vspace{-2pt}\paragraph{Non-triviality condition} 
In both training stages, we implement the \textit{non-triviality condition} in the following steps.
(i)~We select a random \textit{subset} of 20 samples that passed the \textit{informativeness} condition (denoted $X_\text{info}$).
(ii)~From $X_\text{info}$, we iteratively \textit{pick} a sequence of $i \in 1..k$ demonstrations (with $k: 2\leq k \leq 8$) as follows:
\begin{enumerate}
    \vspace{-3pt}\item For each sample $(x_j, y_j) \in X_\text{info}$, we use the training model to compute a \textit{likelihood} of generating the correct prediction $y_\text{pred}$ if a given sample $(x_j, y_j)$ is included among demonstrations. Whenever $y_\text{pred}$ contains more than one token, we compute the likelihood as the \textit{average} of the likelihoods of \textit{all} $y_\text{pred}$'s tokens in the teacher-forced generation.
    \vspace{-3pt}\item In each step $i$, we pick among the demonstrations a sample with which the likelihood of generating correct prediction is \textit{minimal}.
\end{enumerate}
An overview of this process is depicted in Figure~\ref{fig:coat}, with a schematic example of a training prompt in Figure~\ref{fig:abstract}. Full training prompts that our implementation of CoAT constructs in training on each dataset can be found in Table~\ref{tab:full_examples_train} in the Appendix.

\vspace{-3pt}
\section{Experiments}
\label{sec:experiments}
\vspace{-3pt}

    Our experiments provide empirical evidence towards answering three research questions (\textbf{RQs}):

\begin{enumerate}
    \item \textbf{Can we improve models' ability to \textit{benefit} from new reasoning concepts in-context?}
    \item \textbf{Are the concept-aware in-context learners more \textit{robust} to known functional artifacts?}
    \item \textbf{Can concept-aware in-context learning also improve performance in \textit{real-world} tasks?}
\end{enumerate}

The first two RQs validate our assumptions on concept-aware training: that (1) the implementation of CoAT indeed \textit{improves} models' utilisation of both seen and out-of-distribution concepts from demonstrations, and that (2) such an ability \textit{can} make the in-context learning of a CoAT-trained language model more robust to artefacts revealed in previous in-context learners \cite{wei2023larger}. Finally, in (3), we assess whether the enhanced models' ability to rely more on latent concepts can also improve the practical quality of low-resource in-context learning.

\subsection{Training and Evaluation}
\label{sec:training_evaluation_setup}

To maximise comparability with the previous work, we fine-tune our models from \textsc{mT5} pre-trained models of \citet{xue-etal-2021-mt5}. In both training stages (Sec.~\ref{sec:coat_implementation}), we fine-tune all model parameters in a teacher-forced next-token prediction until convergence of evaluation loss.\footnote{Implementations of CoAT training and concept-learning evaluations are on \url{https://github.com/MIR-MU/CoAT}} We further detail the training parameters in Appendix~\ref{appx:training_details}.

In all experiments, we construct evaluation prompts from $k=3$ demonstrations chosen consistently for all models, with prompts including the options for expected labels. We complement all our evaluations with confidence intervals from the bootstrapped evaluation (population $n=100$, repeats $r=200$). We specify evaluation setup separately for each experiment (§\ref{sec:rq1}--\ref{sec:rq3}) with further details and examples in Appendix~\ref{appx:evaluation}.

\subsection{Baselines}
\label{sec:baselines}

We assess the impact \textsc{CoAT}'s main design choices against two baselines, allowing us to measure the impact of both its data construction conditions.

\paragraph{Random demonstrations selection (\textsc{Tk-random})}
We evaluate the impact of all \textsc{CoAT}'s components against a baseline trained in identical settings but picking the in-context demonstrations \textit{randomly} with uniform probability over the whole training set. 
This baseline reproduces the methodology of a majority of the referenced work on instruction tuning, including \textsc{Tk-Instruct} \cite{wang-etal-2022-super} and \textsc{Flan} \cite{chung2022_flan}.
Apart from the demonstration selection, all other settings, including training data, remain identical (§\ref{sec:training_evaluation_setup}) to assure comparability with \textsc{CoAT} models.

\paragraph{Demonstrations passing only \textit{informativeness} condition (\textsc{Tk-info})}
In this baseline, we perform ablation of \textsc{CoAT}'s \textit{non-triviality} condition (Sec.~\ref{sec:coat}) by picking the demonstrations passing \textit{only} the \textit{informativeness} condition. Hence, such-picked demonstrations in the training instructions are informative for the prediction but can exhibit cases where some of the demonstrations are similar or even identical to the predicted sample, making it trivial for the model to perform correct prediction. All other training settings are unchanged (§\ref{sec:training_evaluation_setup}).

\subsection{Other evaluated models}
\label{sec:other_models}

We also evaluate three recent in-context learners for which we can assess which datasets were used in their training mix: (1)~\textbf{\textsc{T0}} of \citet{sanh2022multitask} trained on a mixture of 35 datasets of different tasks in zero-shot settings, mostly of QA type, mapped into a self-containing human-understandable interaction format; (2)~\textbf{\textsc{Tk-Instruct}} of \citet{wang-etal-2022-super} pre-trained in a few-shot format similar to ours, on a mixture of 1,616 diverse tasks, and (3)~\textbf{\textsc{Flan}} models of \citet{chung2022_flan} that further extend data settings of \textsc{Tk-Instruct} to a total of 1,836 tasks, including chain-of-thought labels, i.e. a step-by-step reasoning chain mapping input prompt to a label.

All these models are based on the same pre-trained model (T5), making the results comparable to the level of fine-tuning methodology.
\textsc{Tk-Instruct} and \textsc{Flan} use the data construction reproduced in our \textsc{Tk-random} baseline, but applied in vastly larger data settings.


\vspace{-2pt}
\subsection{\textsc{CoAT}'s ability to improve models utilisation of latent concepts (RQ1)}
\label{sec:rq1}
\vspace{-2pt}

We pose that if the model can utilize a new reasoning concept $\mathcal{C}$ from demonstrations, it will be able to \textit{improve} the prediction in cases where the demonstrations use the same $\mathcal{C}$ as the predicted sample.
Thus, to evaluate if training with \textsc{CoAT} improves models' utilisation of concepts, we evaluate models' performance in a few-shot setting where we ensure that the demonstrations \textit{share} a specific latent concept with the predicted sample. Then, we quantify models' ability to \textit{improve} from the concept by computing the \textit{difference} in accuracy between such concept-sharing evaluation and conventional evaluation using \textit{randomly} chosen demonstrations. 

We perform the first analysis on TeaBReaC with annotated \textit{reasoning chains} as concepts $\mathcal{C}$ shared between demonstrations and predicted sample (Fig.~\ref{fig:abstract}), but to evaluate generalization to \textit{unseen} concepts, we filter out all samples with reasoning chains that were present in training. This results in 316~evaluation scenarios presenting models with 14~previously unseen reasoning patterns.
In this setting, we compare the concept-improving ability of \textsc{CoAT} models with \textsc{Tk-random} baseline.

The important limitation of evaluation with TeaBReaC's concepts is that it remains unclear whether evaluation with \textit{synthetic} samples is representative for concept learning in a \textit{natural} language. Hence, we also perform this analysis with samples and concepts of natural-language tasks.

Previous work of \citet{stefanik2023incontext} evaluated ICL ability over four different functional concepts, all extracted from \textit{explanations} of natural-language datasets. 
We adopt the concepts of this work and evaluate models for in-context learning of the following concepts: 
(i)~\textit{reasoning logic} of NLI samples of GLUE-Diagnostic dataset \cite{wang-etal-2018-glue}, 
(ii)~\textit{entity relations} annotated in human explanations \cite{inoue-etal-2020-r4c} in the HotpotQA dataset \cite{yang-etal-2018-hotpotqa}, 
(iii)~\textit{functional operations} annotated in general elementary-grade tests of OpenBookQA \cite{mihaylov-etal-2018-suit}, 
and (iv)~shared \textit{facts} in science exams of WorldTree dataset \cite{jansen-etal-2018-worldtree,xie-etal-2020-worldtree}. Examples of prompts with concept-sharing demonstrations for these datasets are shown in Table~\ref{tab:full_examples_eval}.

Identically to the case of synthetic concepts, we evaluate the ability of \textsc{CoAT} models to benefit from these concepts when exhibited in demonstrations and compare to uncontrolled demonstrations' selection (\textsc{Tk-random}) used in previous work.

\begin{figure}[tb]
    \vspace{-2pt}
    \centerline{%
     \!\includegraphics[width=0.53\columnwidth]{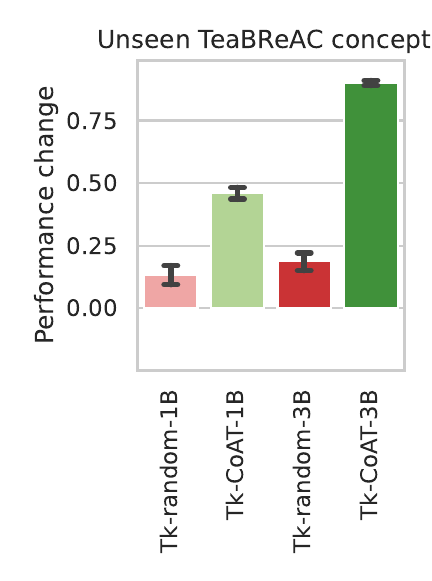}
     \!\!\includegraphics[width=0.53\columnwidth]{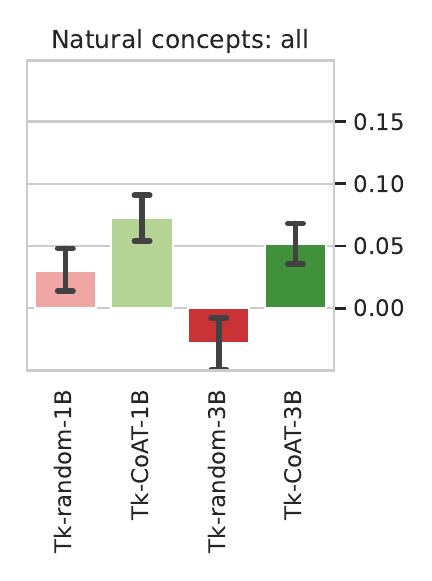}}
     \vspace{-8pt}
    \caption{\textbf{In-context learning of new concepts}: Relative change of performance of models when presented with demonstrations exhibiting a reasoning concept informative for prediction. Evaluation with (left) synthetic TeaBReaC samples, and (right) diverse concepts of \textit{natural} datasets (§\ref{sec:rq1}). 
    }
    \label{fig:concept_learning_avg}
\end{figure}

\subsubsection*{Results}
\paragraph{Concept-aware training improves the ability to benefit from unseen concepts} Figure~\ref{fig:concept_learning_avg} evaluates models' ability to \textit{improve} from presented concepts as the relative difference in performance between random and concept-sharing demonstration selection. First, evaluation with unseen TeaBReaC concepts (left) assesses models' ability to extrapolate the utilisation of latent concepts to 14~previously unseen reasoning chains.
Both \textsc{CoAT} and random-demonstration models (§\ref{sec:baselines}) can improve from concepts presented in demonstrations. However, the improvement of \textsc{CoAT}-trained models is significantly larger and exceeds gains of \textsc{Tk-random} by 2-fold and 4-fold with the smaller and larger model, respectively. This comparison verifies that \textsc{CoAT}'s data construction really improves our targeted skill of better utilizing concepts of demonstrations.

\vspace{-2pt}
\paragraph{CoAT applied with synthetic data also improves the use of \textit{natural} concepts}
\label{parahraph:analysis_natural_concepts}

Evaluation of improvements on selected natural concepts (Figure~\ref{fig:concept_learning_avg}; right) shows that concept-learning ability obtained with synthetic TeaBReaC concepts \textit{transfers} to natural-language settings, as the \textsc{CoAT}-trained models can benefit from concepts significantly \textit{more} than models trained without concept-aware data construction (\textsc{Tk-random}).

Despite that, evaluations over the individual reasoning concepts (Figure~\ref{fig:concept_learning_per_dataset} in Appendix~\ref{appx:per-concept_evaluation}) reveal that even \textsc{CoAT} models can not benefit robustly from \textit{all} concepts. Nevertheless, we note that in the cases where \textsc{CoAT} models do not improve, also \textit{none} of the baselines benefit from presented concepts. This might be attributed to several reasons: (i) the presented concepts are not really \textit{informative} for prediction, (ii)~our training data allowed the models to \textit{memorize} relevant knowledge and, hence, do not \textit{need} (and \textit{benefit from}) the concepts' exposition, or (iii)~our training concepts were simply not sufficient to generalize over these new concepts.

\vspace{-2pt}
\subsection{Robustness of concept-aware in-context learners (RQ2)}
\label{sec:rq2}
\vspace{-2pt}
As we overviewed in Section~\ref{sec:background}, other work reports functional deficiencies of previous in-context learners, including surprising insensitivity of in-context learners to the assigned demonstrations' labels \cite{what-makes-incontext-work}. \citet{wei2023larger} attribute this to models' over-reliance on the \textit{semantic priors} obtained in pre-training, which may override learning of the \textit{functional} concepts. Such behaviour is defective, because the ability to learn functional relations is necessary for robust and interpretable in-context learning of truly unseen tasks.


To evaluate the impact of concept-aware training on models' reliance on their semantic priors, we follow the setup of \citet{wei2023larger} and assess reliance on \textit{labels}' semantics in a standard few-shot evaluation (§\ref{sec:training_evaluation_setup}), with one of the two modifications; (i)~We change the labels to tokens with \textit{irrelevant} meaning for the prediction task, such as `Foo', `Bar' etc. (ii)~We \textit{shuffle} the labels so that semantically incorrect labels are assigned in the demonstrations, but the input-label mapping remains consistent. In both settings, the task's functional relation can still be recovered from demonstrations, but the sole reliance on semantics will either not help or will mislead the model.

In this setting, we evaluate three model types: (i)~\textsc{CoAT}-trained models, (ii)~models with uncontrolled data construction (\textsc{Tk-random} \& previous work), and (iii)~models with uncontrolled data construction, but fine-tuned \textit{only} on a \textit{natural} QA dataset (denoted \textsc{Tk-QA}).
We perform the evaluation over 8~SuperGLUE tasks with discrete labels.

\def\hvezda{*}
\def\activehvezda{\rlap{*}}

\begin{figure}[tb]
    \vspace{-5pt}
    \centerline{%
    \hspace{-14pt}
    \includegraphics[width=0.63\columnwidth]{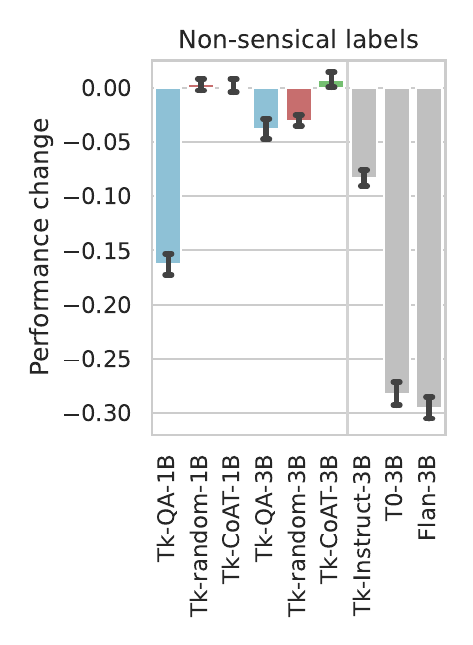} \!\!\!\!\!\includegraphics[width=0.44\columnwidth]{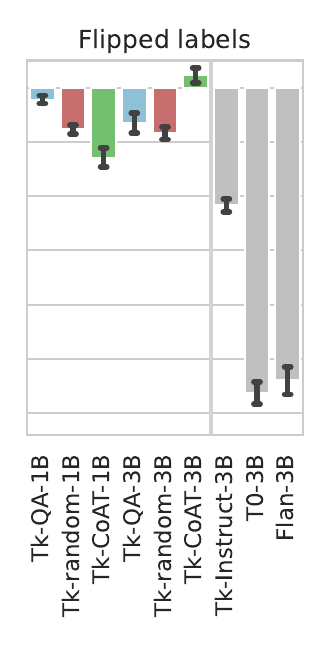}}
    \vspace{-5pt}
    \caption{\textbf{Models' reliance on semantic priors}: Relative change of models' performance when we \textbf{(left)} \textbf{replace} labels with `non-sensical' tokens with no correspondence to the semantics of the task, such as `\textit{foo}', `\textit{bar}', etc.; and \textbf{(right)} \textbf{flip} the original labels, so that e.g. `\textit{negative}' label corresponds to a positive-sentiment sample. \textsc{CoAT} models can in-context learn the input-output mapping similarly well with non-sensical labels and rely on the labels' semantics significantly less than previous in-context learners (in grey).
    }
    \label{fig:semantic_priors}
\end{figure}

\paragraph{Results}
\vspace{-4pt}

Figure~\ref{fig:semantic_priors} shows the results. Evaluation with non-sensical labels (left) reveals that all models pre-trained on a synthetic TeaBReaC dataset (\textsc{Tk-random}, and \textsc{Tk-CoAT}) are more robust to the labels' semantics than our natural-language baseline (\textsc{Tk-QA}). However, a comparison of \textsc{Tk-random} and \textsc{Tk-CoAT} suggests that \textsc{Tk-CoAT}'s preference for learning functional relations is a composition of \textit{both} using a synthetic dataset in pre-training \textit{and} \textsc{CoAT}'s data construction.

A comparison to previous models reveals that all multitask models experience substantially larger decay in performance than our models.
We suspect this feature could be a \textit{bias} specific to massive multi-task learning emerging when label semantics can \textit{explain} a large portion of training data.
This result is consistent with \citet{wei2023larger}, but contrary to their conclusions, we show that ICL robust to semantic distractions does \textit{not} emerge \textit{exclusively} with very large ($\geq100$B) model scale.

Nevertheless, we note that the smaller \textsc{CoAT} model still relies on labels' semantics when recognizable (\textit{Flipped labels}; Fig.~\ref{fig:semantic_priors} right), less than previous models, but comparable to our baselines. 

\begin{table*}[th]
\vspace*{-16pt}
\centering
\tabcolsep2.5dd
\centerline{\scalebox{0.89}{%
\begin{tabular}{@{}lllllllllll@{}}
\hline
               &   
  \multicolumn{1}{c}{AxG} &
  \multicolumn{1}{c}{Ax-b} &
  \multicolumn{1}{c}{WSC} &
  \multicolumn{1}{c}{CB} &
  \multicolumn{1}{c}{RTE} &
  \multicolumn{1}{c}{WiC} &
  \multicolumn{1}{c}{ReCoRD} &
  \multicolumn{1}{c}{BoolQ} &
  \multicolumn{1}{c}{COPA} &
  \multicolumn{1}{c@{}}{MultiRC}\\ \hline
\rowcolor[HTML]{ECF4FF} 
\textsc{Tk-random}-1B   & 49.4\small{±5.2} & 43.6\small{±4.8} & 52.7\small{±5.1} & 21.8\small{±3.9} & 29.3\small{±4.6} & 18.0\small{±4.0} & 15.3\small{±3.8} & 34.0\small{±5.0} & 74.7\small{±3.4} & \nic5.1\small{±2.4}
    \\
\textsc{Tk-random}-3B  &  50.2\small{±5.4} & \underline{57.5}\small{±4.8} & 52.0\small{±5.5} & 47.8\small{±5.1} & 48.9\small{±4.8} & 50.1\small{±4.4} & 16.3\small{±7.3} & 62.8\small{±4.6} & 75.5\small{±2.8} & \nic2.1\small{±1.5}   
    \\ \addlinespace
\rowcolor[HTML]{ECF4FF} 
\textsc{Tk-info}-1B     & 50.0\small{±4.2} & 42.6\small{±5.7} & 52.0\small{±4.3} & \underline{47.2}\small{±3.9 } & 49.2\small{±4.8 } & 53.2\small{±4.5 } & 15.5\small{±4.0  } & 19.6\small{±2.3 } & 61.5\small{±2.3 } & \nic3.2\small{±1.2}
    \\
\textsc{Tk-info}-3B     & 50.8\small{±4.6} & 57.2\small{±4.9} & 53.5\small{±4.8} & 47.3\small{±5.4} & \underline{54.7}\small{±4.9} & 53.6\small{±4.7} & 22.6\small{±4.5} & \underline{64.4}\small{±4.8} & 76.3\small{±3.0} & \nic2.7\small{±2.1}  
    \\ \addlinespace
\rowcolor[HTML]{ECF4FF} 
\textsc{Tk-CoAT}-1B     & \underline{50.4}\small{±5.3} & \underline{52.7}\small{±4.6} & \underline{53.6}\small{±5.2} & 46.9\small{±4.9} & \underline{53.7}\small{±4.9} & \underline{53.5}\small{±5.3} & \underline{17.0}\small{±3.5} & \underline{63.8}\small{±5.4} & \underline{76.1}\small{±3.2} & \underline{11.4}\small{±2.6}
    \\
\textsc{Tk-CoAT}-3B     & \underline{57.9}\small{±4.9} & 57.2\small{±4.8} & \underline{53.6}\small{±4.5} & \underline{60.4}\small{±4.8} & 52.0\small{±5.4} & \underline{56.9}\small{±5.0} & \underline{23.1}\small{±3.8} & 63.6\small{±4.3} & \underline{81.3}\small{±3.3} & \underline{56.9}\small{±3.6}
    \\ \hline
\end{tabular}}}
\caption{\textbf{Effectiveness of concept-aware training: SuperGLUE:} ROUGE-L scores of ICL models evaluated in few-shot setting on SuperGLUE tasks~\cite{wang2019superglue}, trained using (i)~\textit{random} demonstrations sampling used in previous work, (ii)~\textit{informative} demonstrations sampling (§\ref{sec:baselines}) and (iii)~\textit{informative}+\textit{non-trivial} sampling (\textsc{CoAT}; §\ref{sec:coat}). \underline{Underlined} are the best results per each task and model size. See Table~\ref{tab:superglue_others} for a comparison to previous models.
}
\label{tab:coat_ablation}
\end{table*}

\begin{figure}[tb]
         
     \hspace{-5pt}\includegraphics[width=1.07\columnwidth]{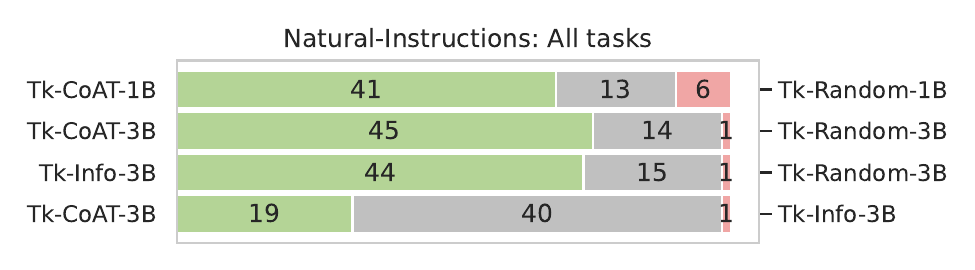}
      
     \vspace{-7pt}\hspace{-5pt}\includegraphics[width=1.07\columnwidth]{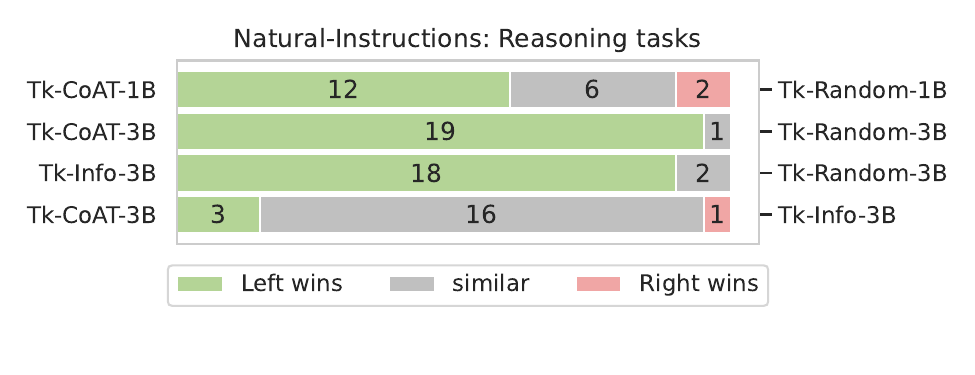}
    \vspace{-25pt}
    \caption{\textbf{Effectiveness of Concept-aware training: Natural-Instructions: } Win rate of models utilising Concept-aware training (\textsc{CoAT}; §\ref{sec:coat}) and traditional instruction tuning (\textsc{Tk-Random}; §\ref{sec:baselines}) evaluated on (top) \textit{all} and (bottom) \textit{reasoning} tasks of Natural-Instructions collection. Values indicate the number of tasks where the referenced model reaches significantly higher accuracy than the other. For the \textit{similar} tasks, the difference in models' performance is not statistically significant.
    }
    \label{fig:coat_ablation_NI}
\end{figure}
\vspace{-2pt}
\subsection{Practical effectiveness of concept-aware in-context learners (RQ3)}
\label{sec:rq3}
\vspace{-2pt}
Finally, we assess the practical quality of concept-aware ICL on previously unseen tasks in a simulated low-resource application with only three \textit{randomly-chosen} demonstrations. We evaluate on samples from two collections of tasks: (i)~SuperGLUE \cite{wang2019superglue} consisting of 10 tasks requiring a variety of reasoning skills, and (ii)~a test split of Natural-Instructions \cite{wang-etal-2022-super} from which we pick 60~extractive tasks.
For SuperGLUE tasks, we verbalize both the demonstrations and predicted sample using all available templates within PromptSource library \cite{bach2022promptsource} and report results for the best-performing template for each model.
For Natural-Instructions tasks, we prefix the demonstrations with the instruction provided with each task.
To maximise evaluation reliability over all models, we analyse the error cases and choose to report the results in ROUGE-L for SuperGLUE, and in a standard accuracy for Natural-Instructions. We specify the metrics selection analysis and other evaluation details in Appendix~\ref{appx:evaluation}, with prompt examples in Table~\ref{tab:full_examples_sglue_NI}.

As a primary reference point, we again compare the results of \textsc{CoAT}-trained models to \textsc{Tk-random}, where we can make sure that all other training configurations except for the data construction method are identical. Further, we compare to \textsc{Tk-info} (without \textit{Non-triviality} condition; §\ref{sec:baselines}) to also evaluate the importance of the \textit{non-triviality} condition. Finally, to provide additional context to our results, we also compare the performance of \textsc{CoAT}-trained models to previous in-context learners (§\ref{sec:other_models}).

\vspace{-3pt}
\paragraph{Results}
Figure~\ref{fig:coat_ablation_NI} compares the accuracy of \textsc{CoAT}-trained models to our baselines: (i) without systematic demonstrations selection (\textsc{Tk-Random}) and (ii) without the \textit{non-triviality} condition (\textsc{Tk-Info}), over 60 tasks of NaturalInstructions collection. In comparison to \textsc{Tk-Random}, \textsc{CoAT} models reach significantly higher accuracy on 41 and 45 of 60 tasks, with comparable performance on a majority (13 and 14) of other tasks. The difference is further magnified on \textit{reasoning} tasks, which we argue might better evaluate models' ability to in-context learn a \textit{functional} relation of the new task. A comparison of \textsc{Tk-Info} with \textsc{Tk-Random} shows that the performance on reasoning tasks is mainly fostered by the \textsc{CoAT}'s \textit{informativeness} condition, but in a full task collection, \textsc{Tk-CoAT} still outperforms \textsc{Tk-Info} in 19 out of 60 tasks. Evaluations on other task segments can be found in Appendix~\ref{appx:NI_different_types}.

In the evaluation over the tasks of SuperGLUE collection (Table~\ref{tab:coat_ablation}), we additionally report the specific values of ROUGE-L that our baselines and CoAT models achieve. With a single exception, models utilising a concept-based selection of demonstrations (\textsc{Tk-CoAT} and \textsc{Tk-Info}) consistently reach higher scores than \textsc{Tk-Random}. Our analyses of models' predictions reveal that in 7 out of 20 evaluations, \textsc{Tk-Random} models fail to follow the task's instruction, consequentially responding out of valid label space. \textsc{Tk-CoAT} is shown to mitigate this issue in all cases except for a smaller \textsc{CoAT}-trained model on MultiRC. A comparison of \textsc{Tk-CoAT} with \textsc{Tk-Info} shows that \textit{non-triviality} condition is more substantial for a smaller model, but the models of both sizes benefit similarly from the concept-sharing selection of demonstrations.

\begin{figure}[tb]
     \hspace{-5pt}\includegraphics[width=1.06\columnwidth]{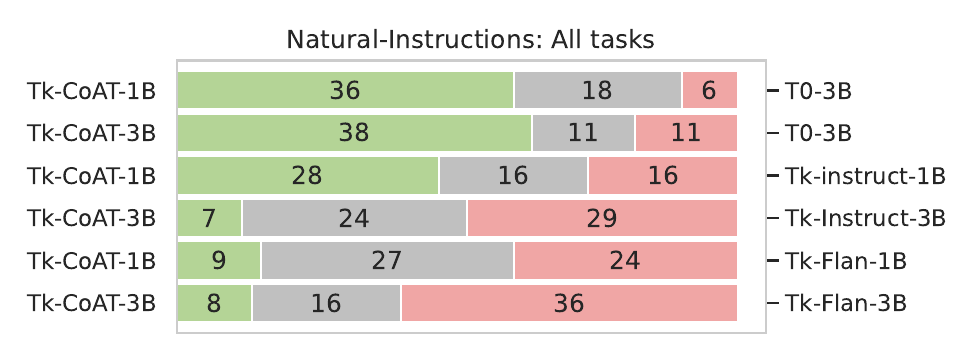}
      
     \vspace{-7.5pt}\hspace{-5pt}\includegraphics[width=1.06\columnwidth]{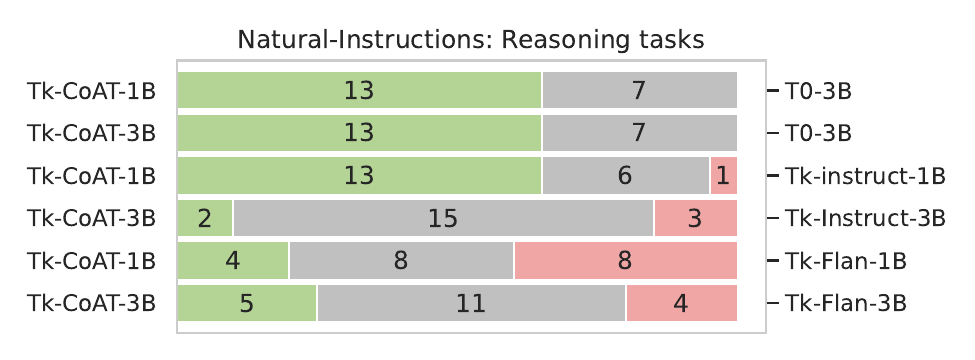}

     \vspace{-7.5pt}\hspace{-5pt}\includegraphics[width=1.06\columnwidth]{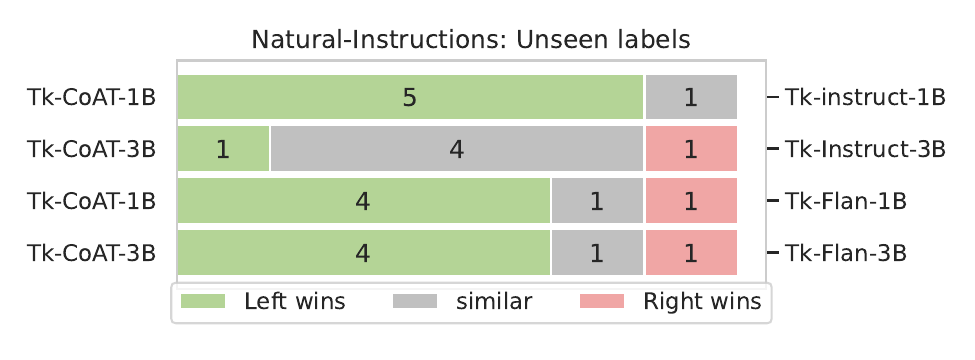}
    \caption{\textbf{Performance comparison to previous work: Natural-Instructions collection:} Win rate of \textsc{CoAT} models trained using two (2) tasks and previous in-context learners trained on mixtures of 35 (\textsc{T0}), 1,616 (\textsc{Tk-Instruct}) and 1,836 tasks (\textsc{Tk-Flan}). Values denote the number of tasks where the model reaches significantly better accuracy. Evaluations over (top) all tasks, (middle) reasoning tasks, (bottom) tasks with labels not present in the training mix of \textsc{Tk-Instruct} and \textsc{Tk-Flan}.
    \vspace{-12pt}
    }
    \label{fig:prev_work_NI}
\end{figure}

\paragraph{Comparison to multitask learners}

Figure~\ref{fig:prev_work_NI} contextualizes the performance of \textsc{CoAT} models trained on two datasets of a single (QA) task with existing instructional models trained on massive mixtures of 35--1,836 tasks. Over all the NI tasks (Fig.~\ref{fig:prev_work_NI}; \textit{top}), \textsc{CoAT} models outperform multitask learners on a majority of tasks in 3 of 6 competitions. CoAT models are outperformed by \textsc{Flan} models but perform at least \textit{similarly} for the \textit{majority} of the tasks in 5 out of 6 competitions. 
The evaluation on reasoning tasks (Fig.~\ref{fig:prev_work_NI}; \textit{middle}) supports our hypothesis that \textsc{CoAT} particularly promotes improvements in in-context learning of new \textit{reasoning} abilities, winning on reasoning tasks over \textsc{Flan} and \textsc{Tk-Instruct} in a comparable number of cases than the opponents. 

Finally, we look at a few tasks with \textit{unseen labels} for both \textsc{Tk-Instruct} and \textsc{Flan} models (Fig.~\ref{fig:prev_work_NI}; \textit{bottom}) where multitask learners can not rely on shortcuts based on unseen tasks' labels. Here, the results of competition between \textsc{CoAT} with \textsc{Flan} models \textit{turns over}, with \textsc{CoAT} models performing significantly better on 4 out of 6 tasks. While this sole segment is not big enough for robust conclusions, the results further support our claim that concept-based ICL is \textit{more robust} to semantic distractions (RQ2).

Table~\ref{tab:superglue_others} in Appendix~\ref{apx:detailed_evaluations} details models' scores on SuperGLUE tasks, providing further evidence on overall comparability of \textsc{CoAT} models to multitask learners. For instance, a comparison with \textsc{Tk-Instruct} reveals that \textsc{CoAT}'s 1B and 3B models reach higher absolute scores on 3 and 5 out of the 7 \textsc{Tk-Instruct}'s unseen tasks.

\section{Conclusion}
\label{sec:conclusion}

Inspired by data-centric theories on emergence of in-context learning, we propose and implement a Concept-aware Training framework for constructing training scenarios that challenge language models to learn to \textit{utilise} latent concepts from in-context prompts. We show that language models \textit{can} be trained to benefit from \textit{unseen} concepts (RQ1), and that such ICL \textit{is} more robust in learning \textit{functional} relations of a new task from demonstrations (RQ2). Finally, in extrinsic evaluation over 70 tasks, we demonstrate the practical efficiency of concept-dependent training data, with \textsc{CoAT} models bringing significant improvements on 41 and 45 out of 60 Natural-Instructions tasks, or 6 and 5 of 10 SuperGLUE tasks (RQ3), while reaching a performance comparable to multitask learning requiring \textit{magnitudes} of more data.

More broadly, our work pioneers an alternative direction for scaling the quality of in-context learning to the previously explored \textit{model} and \textit{data scale} axes. 
We wish towards inspire future work to a more proactive approach to refining training data properties so that fitting such data \textit{necessitates} the emergence of the specific, robust abilities of the models, such as the concept-learning ability.

Specifically, future work can build upon our findings in researching ways to upscale concept-dependent data in unsupervised settings, allowing for pre-training more robust language models with a fraction of data and computing budget.


\section*{Limitations}
\label{sec:limitations}

Although our main objective is to assess the efficiency of concept-aware training, we acknowledge the limitations of our comparison to the previous work, where several aspects convolute the representative comparison of different in-context learners: (i) each of the multitask learners was trained on a different, yet massive set of tasks, making it difficult to find a broader collection that is \textit{new} for multiple models; For this purpose, we surveyed three standard collections used for few-shot evaluation: CLUES \cite{cluesteam2021}, RAFT \cite{alex2021raft} and FLEX \cite{Bragg2021FLEXUE}, but found in total only three tasks unseen by the multitask learners of previous work, all of the same type (classification). Therefore, in our evaluations, we use (a) Tk-Instruct's own evaluation set and (b) SuperGLUE, which significantly overlaps with the training tasks of previous work. (ii) many aspects make it ``easier'' for the model to improve, including the domain of labels or prompt format matching the training distribution (relevant to \textsc{Tk-Instruct} and \textsc{Flan} evaluated on Natural-Instructions). 

Another aspect that we neglect in our experiments in favour of more in-depth analyses is the \textit{impact of pretraining} projected into the properties of the foundation model that we use. We pick \textsc{T5} as a base model to maximise comparability with previous work. While we do not identify any concrete reason to assume that \textsc{CoAT} would perform worse with other base models, one should note that our results do not provide any evidence in this respect.

Finally, we note that the applicability of \textsc{CoAT} is conditioned by the availability of the annotated \textit{concepts} $\mathcal{C}$ in the training datasets, which might be difficult to obtain for natural-language datasets. Our implementation circumvents this issue by using a synthetically curated dataset. Hence, we simultaneously show that concept-aware abilities can also be obtained in the restrictive settings of synthetic-dataset pre-training, where we note that the volume and variability of the synthetic dataset can be scaled further much easier than the natural dataset(s) \cite{trivedi-etal-2022-teaching}. Nevertheless, our experiments do not provide any empirical evidence for answering \textit{to what extent} could further extension of synthetically-generated datasets, possibly covering even more complex concepts, \textit{scale} to further performance gains.

\section*{Ethical Considerations \& Broader Impact}
\label{sec:ethical}

The primary motivation of our work is to minimise the computing demands for the creation of accurate in-context learners by deepening our understanding of the covariates of the resulting quality. We believe that our presented method, as well as the future data-efficient methods improving our understanding of in-context learning, will enable the democratization of the creation of robust and accurate in-context learning models for both research and industry.

Finally, we note that data-efficient methods for training ICLs (as opposed to \textit{multitask training}) might open possibilities for creating more accurate ICLs specialized to languages outside English, where training datasets are scarce. We look forward for the future work that will explore the potential of data-efficient instruction tuning specifically on the target-language datasets, creating in-context learners specially tailored for target languages outside English.

\bibliography{marekov,stefanik}
\bibliographystyle{acl_natbib}

\appendix

\section{Training details}
\label{appx:training_details}

Table~\ref{tab:full_examples_train} shows a full training example for each stage of training: (1)~TeaBReaC with synthetic contexts (top) and (2)~AdversarialQA with natural-language contexts (bottom). In all our training setups, we fine-tune all model parameters for teacher-forced next-token prediction, conventionally used in training sequence-to-sequence language models. In the two training stages (TeaBReaC and AdversarialQA), we use a \textbf{learning rate} of $5e^{-5}$ and $2e^{-5}$, respectively. Other parameters remain identical between stages: effective \textbf{batch size} = 30~samples and \textbf{early stopping} with the patience of 2,000 updates based on evaluation loss on a standardized validation set of each dataset. We do not report the absolute values of evaluation loss as these are not directly comparable. In \textsc{CoAT} training, we use a random subsample of $20$~informative examples as a candidate set for a selection of non-trivial demonstrations.

Other parameters of training configuration default to Training Arguments of Transformers library \cite{Wolf2019HuggingFacesTS} in version 4.19.1.
For readability, we implement the relatively complex demonstrations' selection as a new objective of the Adaptor library~\cite{stefanik-etal-2022-adaptor}.
The picked demonstrations are encoded into a format consistent with the evaluation.

\section{Evaluation details}
\label{appx:evaluation}

Tables~\ref{tab:full_examples_eval} shows an example of an instruction for each evaluation that we perform within the concept-learning evaluation. For readability, we only shorten the examples of HotpotQA, where we omit some sources of data available for the model. In the case of TeaBReaC not shown in this table, the evaluation prompt format is the same as in training (Table~\ref{tab:full_examples_train}), whereas we make sure that the reasoning chains of evaluation samples differ from the training.

\begin{table*}
  \centering\tabcolsep 3dd
    \scalebox{0.7}{%
  \begin{tabularx}{1.428\textwidth}{@{} l >{\raggedright\arraybackslash} p{30mm} p{157mm} >{\raggedright\arraybackslash}X r@{}}
    Dataset&Concept&Training instruction&Target \\
    \midrule
        TeaBReaC & 
        \textbf{Exactly-matching reasoning chain} $[$\textit{"select" $\rightarrow$ "maximum" $\rightarrow$ "list" $\rightarrow$ "maximum" $\rightarrow$ "sum"$]$} & 
        ``\textbf{Input:} how many points did the Monte Vesuvio" score in their two highest scoring matches? \textbf{Context:} scores of games of Pentagon". 99 scores of games of monte vesuvio". 67 scores of games of Pentagon". 6 scores of games of monte vesuvio". 76 scores of games of Pentagon". 37 scores of games of monte vesuvio". 56 scores of games of Pentagon". 8 scores of games of Pentagon". 90 scores of games of Pentagon". 20 Answer: \textbf{Prediction:} 143 \textbf{[2 more examples]} \textbf{Input:} how many points did the Bell 212 score in their two highest scoring games? \textbf{Context:} scores of games of bell 212. 90 scores of games of S-50. 54 scores of games of bell 212. 41 scores of games of bell 212. 36 scores of games of S-50. 23 scores of games of bell 212. 6 scores of games of bell 212. 2 scores of games of S-50. \textbf{Prediction:} '' & 
        ``131'' 
    \\ \addlinespace
        AdversarialQA &
        \textbf{Matching question-word} ``Who'' &
        ``\textbf{Input:} Who was the Speaker in 1909? \textbf{Context:} Second, Democrats have always elevated their minority floor leader to the speakership upon reclaiming majority status. Republicans have not always followed this leadership succession pattern. In 1919, for instance, Republicans bypassed James R. Mann, R-IL, who had been minority leader for eight years, and elected Frederick Gillett, R-MA, to be Speaker. Mann "had angered many Republicans by objecting to their private bills on the floor;" also he was a protégé of autocratic Speaker Joseph Cannon, R-IL (1903–1911), and many Members "suspected that he would try to re-centralize power in his hands if elected Speaker." More recently, although Robert H. Michel was the Minority Leader in 1994 when the Republicans regained control of the House in the 1994 midterm elections, he had already announced his retirement and had little or no involvement in the campaign, including the Contract with America which was unveiled six weeks before voting day. \textbf{Prediction:} Joseph Cannon, R-IL. \textbf{\textbf{[2 more examples]}} \textbf{Input:} Who created the legal system still in use in Florida? \textbf{Context:} As a result of these initiatives northeastern Florida prospered economically in a way it never did under Spanish rule. Furthermore, the British governors were directed to call general assemblies as soon as possible in order to make laws for the Floridas and in the meantime they were, with the advice of councils, to establish courts. This would be the first introduction of much of the English-derived legal system which Florida still has today including trial by jury, habeas corpus and county-based government. Neither East Florida nor West Florida would send any representatives to Philadelphia to draft the Declaration of Independence. Florida would remain a Loyalist stronghold for the duration of the American Revolution. \textbf{Prediction:} '' &
        ``British''
    \\ \bottomrule        
        
  \end{tabularx}
}
  \caption{Examples of \textbf{training instructions} with expected outputs, for both our datasets applied in training. Note that the shared reasoning concept is not a part of the model's input.}
  \label{tab:full_examples_train}
\end{table*}

\begin{table*}
  \centering\tabcolsep 3dd
    \scalebox{0.7}{%
  \begin{tabularx}{1.429\textwidth}{@{} l >{\raggedright\arraybackslash} X p{135mm} >{\raggedright\arraybackslash} p{19mm}@{} }
    Dataset&Concept&Model instruction&Expected output \\
    \midrule
        GLUE NLI Diag. &
        Double negation &
        ``\textbf{Input:} I will say that she stole my money. Question: I won't say that she didn't steal my money. True, False, or Neither? \textbf{Prediction:} Neither \textbf{Input:} I won't say that she didn't steal my money. Question: I will say that she stole my money. True, False, or Neither? \textbf{Prediction:} Neither \textbf{Input:} A rabbi is at this wedding, standing right there standing behind that tree. Question: It's not the case that there is no rabbi at this wedding; he is right there standing behind that tree. True, False, or Neither? \textbf{Prediction:} True \textbf{Input:} Even after now finding out that it's animal feed, I won't ever stop being addicted to Flamin' Hot Cheetos. Question: Even after now finding out that it's animal feed, I will never stop being addicted to Flamin' Hot Cheetos. True, False, or Neither? \textbf{Prediction:} ''&
        ``True''
    \\ \addlinespace
        OpenBookQA &
        \textbf{Shared facts:} \{"Earth is  greater in mass than Mars", "gravity means gravitational pull; gravitational force; gravitational attraction", "as the force of gravity increases, the weight of objects will increase."\} &
        ``\textbf{Facts:} a decrease is a kind of change. increase means more. as mass of a planet; of a celestial body increases, the force of gravity on that planet will increase. to change means to become different. an animal is a kind of living thing. the gravitational force of a planet; of a celestial object does not change the mass of an object on that planet or celestial body. an increase is the opposite of a decrease. an astronaut is a kind of human. massive means great in mass. the Mars Rover is a kind of vehicle. a living thing is a kind of object. Earth is greater in mass than Mars. gravity means gravitational pull; gravitational energy; gravitational force; gravitational attraction. greater means higher; more in value. stay the same means not changing. a moon is a kind of celestial object; body. an increase is a kind of change. Earth is a kind of planet. as the force of gravity increases, the weight of objects will increase. less is similar to decrease. Mars is a kind of planet. \textbf{Input:} An object has a weight of 10 kg on the surface of Earth. If the same object were transported to the surface of Mars, the object would have a weight of 3.8 kg. Which best explains why the weight of the object changed when transported from Earth to Mars? (A) The density of the object is greater on Earth than it is on Mars. (B) The volume of the object is greater on Earth than it is on Mars. (C) Gravitational force is greater on Earth than it is on Mars. (D) Atmospheric pressure is less on Earth than it is on Mars. \textbf{Prediction:} Gravitational force is greater on Earth than it is on Mars \textbf{[two more examples]} \textbf{Input:} When astronauts walked on the Moon, they used weighted boots to help them walk due to the lower gravitational pull. What difference between Earth and the Moon accounts for the difference in gravity? (A) density (B) diameter (C) mass (D) volume. \textbf{Prediction:} '' &
        ``mass''
    \\ \addlinespace
        HotpotQA &
        \textbf{Shared relation in reasoning}: ``X is a genus'' &
        ``\textbf{Input:} Are Broughtonia and Laeliocattleya both orchids? Hint: use the information from the paragraphs below to answer the question. Otaara, abbreviated Otr. in the horticultural trade, is an intergeneric hybrid of orchids, with "Brassavola", "Broughtonia", "Cattleya", "Laelia" and "Sophronitis" as parent genera.  Paracaleana commonly known as duck orchids, is a genus of flowering plants in the orchid family, Orchidaceae that is found in Australia and New Zealand. Duck orchids have a single leaf and one or a few, dull-coloured, inconspicuous flowers. (...) \textbf{Prediction:} yes \textbf{[two more examples]} \textbf{Input:} Are both Parodia and Thalictrum flowering plants? Hint: use the information from the paragraphs below to answer the question. - Thalictrum ( ) is a genus of 120-200 species of herbaceous perennial flowering plants in the Ranunculaceae (buttercup) family native mostly to temperate regions. Meadow-rue is a common name for plants in this genus. - Parodia is a genus of flowering plants in the cactus family Cactaceae, native to the uplands of Argentina, Peru, Bolivia, Brazil, Colombia and Uruguay. This genus has about 50 species, many of which have been transferred from "Eriocactus", "Notocactus" and "Wigginsia". They range from small globose plants to 1 m tall columnar cacti. All are deeply ribbed and spiny, with single flowers at or near the crown. Some species produce offsets at the base. They are popular in cultivation, but must be grown indoors where temperatures fall below 10 degrees. \textbf{Prediction:} ''&
        ``yes''
    \\ \addlinespace
        WorldTree &
        \textbf{Relation of objects}: "generate" &
        ``\textbf{Input:} Despite what some think, instead around themselves, our planet spins around... Choices: pluto, the moon, the milky way, the sun. \textbf{Prediction:} the sun \textbf{Input:} In a single year, a giant globe will do this to a giant star. Choices: fight, burn, circle, explode. \textbf{Prediction:} circle \textbf{Input:} The earth revolves around... Choices: a heat source, the Milky Way, a neighboring planet, the moon. \textbf{Prediction:} a heat source \textbf{Input:} the central object of our solar system is also... Choices: the smallest object in the solar system, the coldest heavenly body, the farthest star from us, the closest star from us. \textbf{Prediction:} ''&
        ``the closest star from us''
    \\ \midrule        
  \end{tabularx}
}
  \caption{Examples of \textbf{evaluation instructions} with expected outputs, for each dataset used in evaluation of in-context learning of new concepts (RQ1). Note that the demonstrations within the instructions share the annotated \textit{Concept} with the following \textit{predicted sample}.}
  \label{tab:full_examples_eval}
\end{table*}

\begin{table*}
  \centering\tabcolsep 3dd
    \scalebox{0.7}{%
  \begin{tabularx}{1.43\textwidth}{@{} l >{\raggedright\arraybackslash} l X >{\raggedright\arraybackslash} p{20mm}@{} }
    Dataset&Concept&Model instruction&Expected output \\
    \midrule
        SuperGLUE &
        - &
        ``\textbf{Input}: The soldiers were concealed in the brush. Select the most plausible  cause: - They were armed with rifles. - They wore camouflage uniforms. \textbf{Prediction}: They wore camouflage uniforms. \textbf{Input}: The print on the brochure was tiny. Select the most plausible  effect: - The man put his glasses on. - The man retrieved a pen from his pocket. \textbf{Prediction}: The man put his glasses on. \textbf{Input}: I excused myself from the group. Select the most plausible  cause: - I turned off my phone. - My phone rang. \textbf{Prediction}: My phone rang. \textbf{Input}: My body cast a shadow over the grass. Select the most plausible  cause: - The sun was rising. - The grass was cut. \textbf{Prediction}:''&
        ``The sun was rising.''
    \\ \addlinespace
        Natural-Instructions &
        - &
        ``Indicate with `Yes` if the given question involves the provided reasoning `Category`. Indicate with `No`, otherwise. We define five categories of temporal reasoning. First: "event duration" which is defined as the understanding of how long events last. For example, "brushing teeth", usually takes few minutes. Second: "transient v. stationary" events. This category is based on the understanding of whether an event will change over time or not. For example, the sentence "he was born in the U.S." contains a stationary event since it will last forever; however, "he is hungry" contains a transient event since it will remain true for a short period of time. Third: "event ordering" which is the understanding of how events are usually ordered in nature. For example, "earning money" usually comes before "spending money". The fourth one is "absolute timepoint". This category deals with the understanding of when events usually happen. For example, "going to school" usually happens during the day (not at 2 A.M). The last category is "frequency" which refers to how often an event is likely to be repeated. For example, "taking showers" typically occurs ~5 times a week, "going to Saturday market" usually happens every few weeks/months, etc. \textbf{Input:} Sentence: Jack played basketball after school, after which he was very tired. Question: How long did Jack play basketball? Category: Event Duration. \textbf{Prediction:} Yes \textbf{Input:} Sentence: He was born in China, so he went to the Embassy to apply for a U.S. Visa. Question: How often does he apply a Visa? Category: Frequency. \textbf{Prediction:} Yes \textbf{Input:} Sentence: Jack played basketball after school, after which he was very tired. Question: Was Jack still tired the next day? Category: Event Duration. \textbf{Prediction:} No \textbf{Input:} Sentence: It refers to a woman who is dangerously attractive, and lures men to their downfall with her sexual attractiveness. Question: How long does it take to lure men to their downfall? Category: Event Duration. \textbf{Prediction:} ''&
        ``Yes''
    \\ \midrule
    
  \end{tabularx}
}
  \caption{Examples of \textbf{evaluation instructions} with expected outputs, for selected tasks of \textbf{SuperGLUE} and \textbf{Natural-Instructions} (RQ3). Displayed samples are from CoPA and MCTato Temporal Reasoning tasks, respectively. Note that in these evaluations, demonstrations are picked \textbf{randomly}, regardless of their concepts.}
  \label{tab:full_examples_sglue_NI}
\end{table*}

\paragraph{SuperGLUE Evaluation format}
As mentioned in Section~\ref{sec:training_evaluation_setup}, we verbalize both the demonstrations and predicted sample using all available templates of PromptSource library \cite{bach2022promptsource}, obtaining prompts for each demonstration prompt $x_i$ and its label $y_i$ in a free-text form. The prompts commonly contain the full-text match of the possible labels as options for the model.

Following the example of \citet{wang-etal-2022-super}, we additionally prepend the demonstrations and labels with keywords ``Input'' and ``Prediction'' and separate demonstrations with new lines. Thus, the resulting input$\rightarrow$output pairs in evaluation take this format:

\bigskip
\begin{center}
    \noindent \textit{``Input: $x_1$ Prediction: $y_1$ <newline>\\
Input: $x_2$ Prediction: $y_2$ <newline>\\
Input: $x_3$ Prediction: $y_3$ <newline>\\
Input: $x_\text{pred}$ Prediction: '' }$\rightarrow$ ``$y_\text{pred}$''
\end{center}
\vspace{3pt}
where demonstrations $(x_i, y_i)$ are picked randomly but consistently between all evaluated models.

\paragraph{Natural-Instructions Evaluation format}
In the evaluations on Natural-Instructions, we closely follow the example of \citet{wang-etal-2022-super} and additionally prepend the sequence of demonstrations with an instruction provided for each task:
\bigskip
\begin{center}
    \noindent \textit{``<task instruction>\ \ \ \ \ \ \ \  <newline>\\
Input: $x_1$ Prediction: $y_1$ <newline>\\
Input: $x_2$ Prediction: $y_2$ <newline>\\
Input: $x_3$ Prediction: $y_3$ <newline>\\
Input: $x_\text{pred}$ Prediction: '' }$\rightarrow$ ``$y_\text{pred}$''
\end{center}
\vspace{3pt}
where the \textit{<task instruction>} contains the instruction as would be given to the annotators of the evaluation task, usually spanning between 3--6 longer sentences. The demonstrations are again picked randomly but consistently between models.

Examples of evaluation prompts for both SuperGLUE and Natural-Instructions can be found in Table~\ref{tab:full_examples_sglue_NI}.

\paragraph{Evaluation metrics selection}

Previous work training in-context few-shot learners is not consistent in the use of evaluation metrics, and the choice usually boils down to either using the exact-match accuracy \cite{sanh2022multitask,chung2022_flan} or ROUGE-L of \citet{lin-2004-rouge} \cite{wang-etal-2022-super}, evaluating the longest common sequence of tokens. We investigate these two options with the aim of not penalising the models for minor discrepancies in the output format (in the accuracy case) but avoiding false positive evaluations in predictions that are obviously incorrect (in the ROUGE case).

Investigation of the models' predictions reveals that the selection of the metric makes a large difference only in the case of \textsc{Tk-Instruct} models, where the situation differs between SuperGLUE and Natural-Instructions, likely due to the character of the evaluation prompts. 

(1) On \textbf{SuperGlue}, e.g. on MultiRC task, for the evaluation prompt: "Does {answer} sound like a valid answer to the question: {question}", \textsc{Tk-Instruct-3B} in our evaluation predicts "Yes." or "Yes it is" (instead of "Yes"), or "No not at all" (instead of "No"), likely due to the resemblance with the format of training outputs. As we do not wish to penalize these cases, we use ROUGE-L over all SuperGLUE evaluations.

(2) In \textbf{Natural-Instructions} evaluation, we find that \textsc{Tk-Instruct} often predicts longer extracts from the input prompt. This is problematic with ROUGE-L in the cases where the extract contains \textit{all} possible answers, such as in the \textsc{Tk-Instruct-1B}'s prediction: ``yes or no'' to the prompt whose instruction ends with ``Please answer in the form of yes or no.''. As we encounter this behaviour in a large portion of Natural-Instructions tasks, we evaluate all models on Natural-Instructions for exact-match accuracy after the normalization of the casing and the removal of non-alphabetic symbols. To make sure that the model is presented with the exact-matching answer option, we exclude from evaluation the tasks where the correct answer is not presented in the task's instruction. The reference to the list of Natural-Instructions evaluation tasks can be found in Appendix~\ref{appx:eval_tasks}.

For the reported evaluations of the Reasoning tasks, we pick from the list of evaluation tasks the ones concerned with the reasoning task by simply matching the tasks with `reasoning' in their name, resulting in the collection of 20 evaluation tasks.
 
\begin{figure}[t]
    \centering
    \hspace*{-3.5mm}
    \includegraphics[width=1.06\columnwidth]{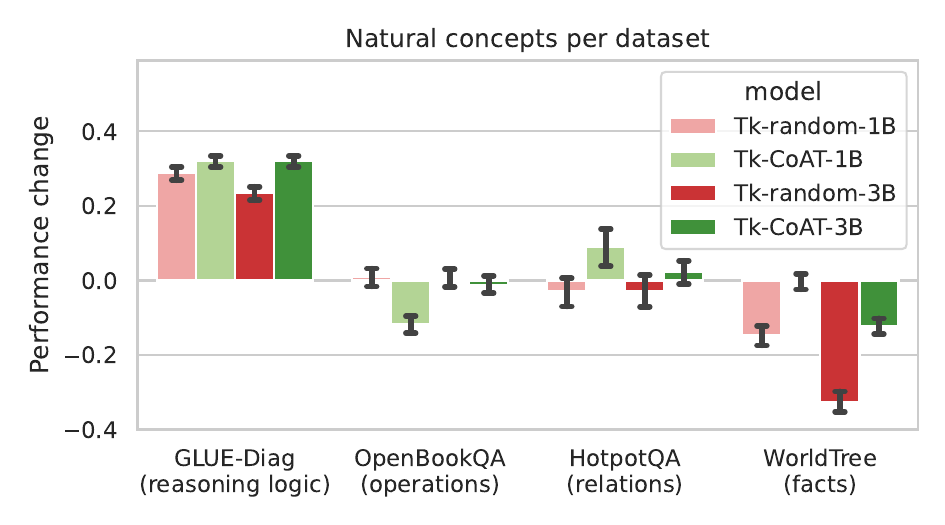}
    \caption{\textbf{In-context learning of natural concepts for each dataset}: While \textsc{CoAT} improves the ability to benefit from reasoning concepts on average (Fig.~\ref{fig:concept_learning_avg}), per-concept evaluation reveals that this ability is not consistently robust.\vspace{-10pt}}
    \label{fig:concept_learning_per_dataset}
\end{figure}

\section{Further evaluations}
\label{apx:detailed_evaluations}

\begin{table*}[bth]
\centering
\tabcolsep2.5dd
\bgroup
\centerline{\scalebox{0.813}{%
\begin{tabular}{@{}lccccccccccc@{}}
\hline
 &
  \multicolumn{1}{l}{\!\!\!\!\!\!\!\!\!\# train tasks} &
  \multicolumn{1}{c}{\!\!\!\!AxG} &
  \multicolumn{1}{c}{Ax-b} &
  \multicolumn{1}{c}{WSC} &
  \multicolumn{1}{c}{CB} &
  \multicolumn{1}{c}{RTE} &
  \multicolumn{1}{c}{WiC} &
  \multicolumn{1}{c}{ReCoRD} &
  \multicolumn{1}{c}{BoolQ} &
  \multicolumn{1}{c}{COPA} &
  \multicolumn{1}{c}{MultiRC} \\ \hline
\rowcolor[HTML]{ECF4FF} 
\textsc{Flan}-1B        & 1,836 & \underline{84.8}\small{±3.9} & 21.9\small{±4.0} & \underline{70.7}\small{±4.8} & 92.5\small{±2.8*} & 92.1\small{±3.0*} & 69.9\small{±5.1*} & 38.9\small{±5.2*} & 92.3\small{±2.7*} & 97.8\small{±1.5*} & 88.3\small{±3.2*}  \\
\textsc{Flan}-3B        & 1,836 & \underline{95.3}\small{±3.7} & 22.0\small{±8.0} & \underline{80.2}\small{±9.2} & 92.7\small{±6.7*} & 96.0\small{±4.0*} & 79.7\small{±8.3*} & 62.2\small{±9.7*} & 92.1\small{±5.1*} & 99.3\small{±1.6*} & 90.4\small{±6.4*} \\ \addlinespace
\rowcolor[HTML]{ECF4FF} 
\textsc{Tk-Instruct}-1B & 1,616 & 51.9\small{±4.9} & \underline{57.2}\small{±5.8} & 49.8\small{±4.9} & 46.0\small{±5.5 } & \underline{55.5}\small{±4.8} & \underline{53.5}\small{±5.3} & 13.1\small{±3.7 } & 63.4\small{±3.4*} & 76.9\small{±3.2*} & 62.2\small{±5.1*} \\
\textsc{Tk-Instruct}-3B & 1,616 & 53.5\small{±4.7} & 49.9\small{±4.9} & 51.2\small{±4.9} & \underline{66.3}\small{±4.6} & \underline{62.7}\small{±4.6} & 50.4\small{±4.8} & 18.6\small{±4.2} & 68.8\small{±4.4*} & 73.8\small{±3.5*} & 59.9\small{±4.9*} \\ \addlinespace
\textsc{T0}-3B          & \hphantom{0,0}35                     & 65.0\small{±4.5} & 36.1\small{±4.6} & 53.5\small{±5.2} & 48.0\small{±5.4 } & 51.3\small{±5.2} & 54.0\small{±5.0 } & 20.5\small{±4.0 } & 60.1\small{±4.9 } & 56.8\small{±3.6 } & 56.2\small{±4.4}  \\ \addlinespace
\addlinespace
\rowcolor[HTML]{ECF4FF} 
\textsc{Tk-CoAT}-1B     & \hphantom{0,00}2     & 50.4\small{±5.3} & 52.7\small{±4.6} & 53.6\small{±5.2} & \underline{46.9}\small{±4.9} & 53.7\small{±4.9} & \underline{53.5}\small{±5.3} & \underline{17.0}\small{±3.5} & \underline{63.8}\small{±5.4} & \underline{76.1}\small{±3.2} & 11.4\small{±2.6}  \\
\textsc{Tk-CoAT}-3B & \hphantom{0,00}2
   & 57.9\small{±4.9} & \underline{57.2}\small{±4.8} & 53.6\small{±4.5} & 60.4\small{±4.8} & 52.0\small{±5.4} & \underline{56.9}\small{±5.0} & \underline{23.1}\small{±3.8} & \underline{63.6}\small{±4.3} & \underline{81.3}\small{±3.3} & \underline{56.9}\small{±3.6} \\ \hline
\end{tabular}}}

\caption{\textbf{ICL performance: comparison to previous ICL models} ROUGE-L of \textsc{CoAT}-trained ICL models and models of comparable size in previous work. Evaluation setup is consistent with Table~\ref{tab:coat_ablation}. In cases marked with $^*$, the task was used in the model's training; \underline{Underlined} are the best results per unseen task and model size.
}
\label{tab:superglue_others}
\egroup
\end{table*}

\subsection{SuperGLUE evaluations of other models}
Table~\ref{tab:superglue_others} compares the performance over the tasks of SuperGLUE collection \cite{wang2019superglue} for \textsc{CoAT} models trained on two tasks of the same (QA) type with in-context learners trained on 35--1,836 tasks of the comparable size. Despite the significantly smaller volumes and complexity of the training dataset, \textsc{CoAT}-trained models show competitive results to similar-size or even larger in-context learners of previous work. For instance, the 1-billion-parameter \textsc{Tk-CoAT} performs better than the 3-billion \textsc{T0} in 3~cases (Ax-b, RTE, COPA) and comparably in another 3 cases (WSC, CB, WiC). In comparison with \textsc{Tk-instruct} of the same size, \textsc{Tk-CoAT-1B} outperforms \textsc{Tk-instruct} in 3 out of 7 unseen tasks (WSC, CB, ReCoRD), and reaches similar scores in most other cases, even in 2 out of 3 tasks that were included in \textsc{Tk-instruct}'s training mix. Similarly, larger \textsc{Tk-CoAT-3B} outperforms \textsc{Tk-instruct} on 4 of 7 new tasks (Ax-b, WSC, WiC, ReCoRD), but with larger gaps on the others.

\subsection{Natural-Instructions: other task types}
\label{appx:NI_different_types}

\begin{figure}[tb]
    \vspace{-8pt}
     \hspace{-5pt}\includegraphics[width=1.06\columnwidth]{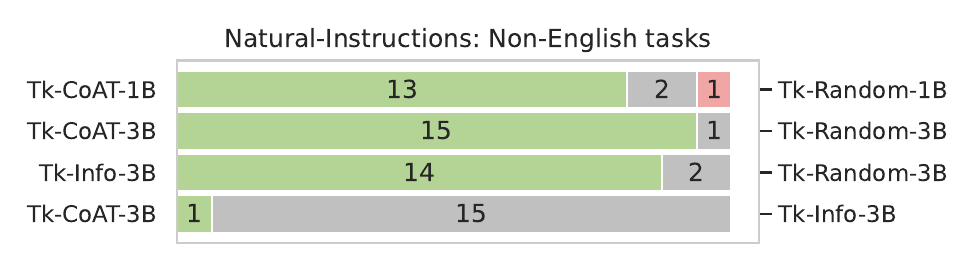}
      
     \vspace{-7.5pt}\hspace{-5pt}\includegraphics[width=1.06\columnwidth]{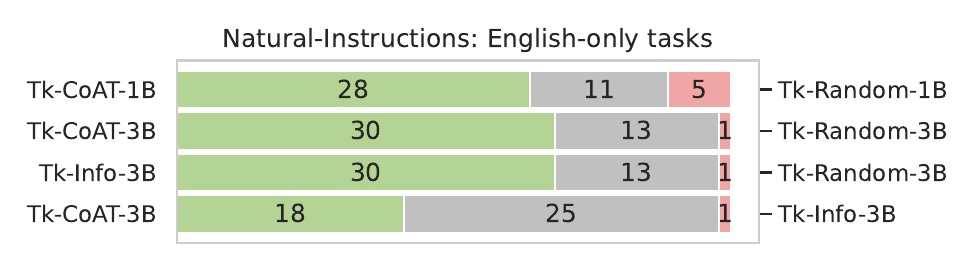}

    \caption{\textbf{Impact of Concept-aware training per different language settings: } Pairwise comparison of models trained using selected training configurations (§\ref{sec:baselines}) on (top) \textit{Non-English} tasks and (bottom) \textit{English-only} tasks of Natural-Instructions collection. Values in green and red bars indicate a number of tasks where the referenced model reaches significantly higher accuracy than the other. For the tasks denoted as \textit{similar}, the difference in performance falls within the evaluation's confidence intervals.}
    \label{fig:our_models_NI_multi}
\end{figure}

Figure~\ref{fig:our_models_NI_multi} evaluates the impact of \textsc{CoAT}'s mechanism on the quality of in-context learning separately on the English and non-English tasks. The figure reveals that \textsc{CoAT} works particularly well for non-English tasks. Our analyses found this is mainly due to the low performance of the baseline on the non-English tasks. We speculate that this can be a consequence of the higher reliance of the baseline on token semantics (Section~\ref{sec:rq3}, RQ2); As our models are fine-tuned on an English-only QA model, such learnt reliance is not applicable in multilingual settings.

\begin{figure}[tb]
    \vspace{-8pt}
     \hspace{-5pt}\includegraphics[width=1.06\columnwidth]{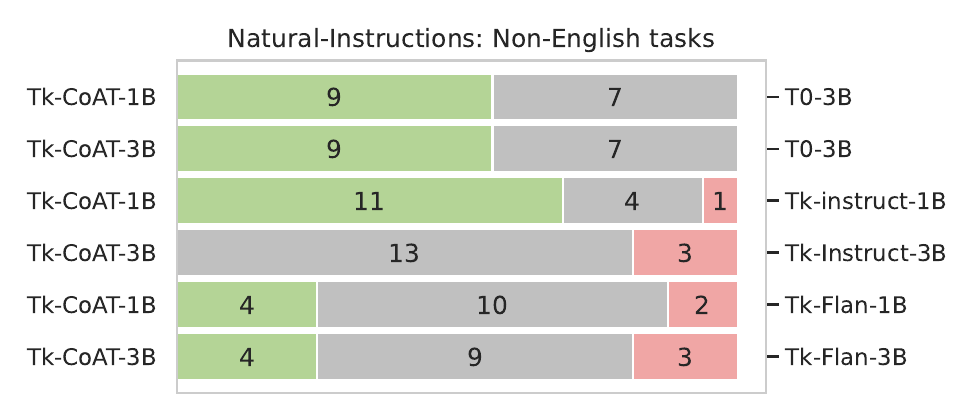}
      
     \vspace{-7.5pt}\hspace{-5pt}\includegraphics[width=1.06\columnwidth]{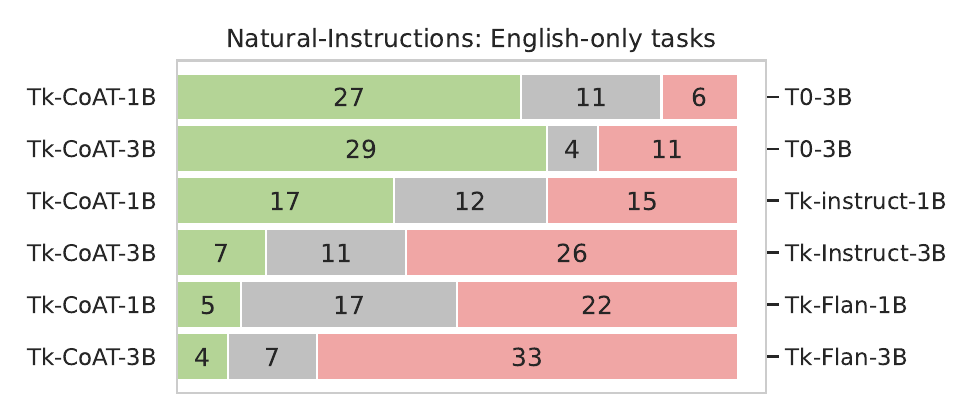}

    \caption{\textbf{Comparison to previous work per different language settings: } Pairwise comparison of \textsc{CoAT} models vs. the models of previous work on (top) \textit{Non-English} tasks and (bottom) \textit{English-only} tasks of Natural-Instructions collection. Values denote the number of tasks where the model reaches significantly better accuracy. For the tasks denoted as \textit{similar}, the difference in performance falls within the evaluation's confidence intervals.}
    \label{fig:prev_work_NI_multi}
\end{figure}

Figure~\ref{fig:prev_work_NI_multi} compares the performance of \textsc{CoAT} models against the models of previous work, separately on the English and non-English tasks. We can see that CoAT is slightly better at the multilingual portion of Natural-Instructions, but the difference is not principal. 

\subsection{Per-concept evaluations}
\label{appx:per-concept_evaluation}

Figure~\ref{fig:concept_learning_per_dataset} evaluates the performance gains of the baseline models (§\ref{sec:baselines}) and \textsc{CoAT}-trained models individually per each of the concepts of the natural datasets. While the \textsc{CoAT} models are able to benefit from concepts the largest in the relative change of quality, they are also not consistent in the ability to benefit from all the concepts. However, as discussed in Section~\ref{parahraph:analysis_natural_concepts}, this does not imply that \textsc{CoAT} is unable to utilize these concepts.

\subsection{Evaluation tasks and other configurations}
\label{appx:eval_tasks}
SuperGLUE \cite{wang2019superglue} consists of the following tasks (as ordered in our Results, §\ref{sec:rq3}): Winogender Schema Diagnostics (AxG) \cite{rudinger-EtAl:2018:N18}, Broadcoverage Diagnostics (CB), The Winograd Schema Challenge, CommitmentBank (CB), Recognizing Textual Entailment (RTE), ContextWords in Context (WiC) \cite{pilehvar-camacho-collados-2019-wic}, Reading Comprehension with Commonsense Reasoning (ReCoRD) \cite{2018arXiv181012885Z}, BoolQ \cite{clark-etal-2019-boolq}, Choice of Plausible Alternatives (COPA), Multi-Sentence Reading Comprehension (MultiRC). 

Natural-Instructions consists of a larger mixture of tasks, which we do not enumerate here to maintain readability; the full list of evaluation tasks can be found in the original work of \citet{wang-etal-2022-super} in Figures~11 and 12.

To maintain comparability of evaluations among models, we deterministically fix the demonstration selection procedure so that only the full prediction prompts for all the models are the same. In the analyses comparing the differences in performance (§\ref{sec:rq1}; RQ1+2), we fixed the prediction samples ($x_\text{pred}$) between different demonstrations' sampling strategies to avoid perplexing our comparison with possible data selection biases. Further details can be found in the referenced implementation.

\section{Computational Requirements}

We run both training and evaluation experiments on a machine with dedicated single \textsc{NVIDIA A100-SXM-80GB}, 40\,GB of RAM and a single CPU core. Hence, all our reproduction scripts can run on this or a similar configuration. Two stages of training in total take at most 6,600 updates and at most 117\,h of training for \textsc{Tk-CoAT} to converge.

\end{document}